\pdfoutput=1

\documentclass[11pt]{article}

\usepackage{naacl2021}

\usepackage{times}
\usepackage{latexsym}

\usepackage[T1]{fontenc}

\usepackage[utf8]{inputenc}

\usepackage{microtype}
\usepackage{float}
\usepackage{tabularx}
\usepackage{adjustbox}
\usepackage{grffile}
\definecolor{darkgreen}{rgb}{0.0, 0.5, 0.0}
\usepackage{booktabs}

\title{MMIU: Dataset for Visual Intent Understanding in Multimodal Assistants}
\author{Alkesh Patel, Joel Ruben Antony Moniz, Roman Nguyen, \\ {\bf Nicholas Tzou, Hadas Kotek, Vincent Renkens}\\
  Apple, Cupertino, CA, USA}
\begin{document}
\maketitle

\begin{figure*}
\renewcommand\thefigure{2}
  \includegraphics[width=\textwidth]{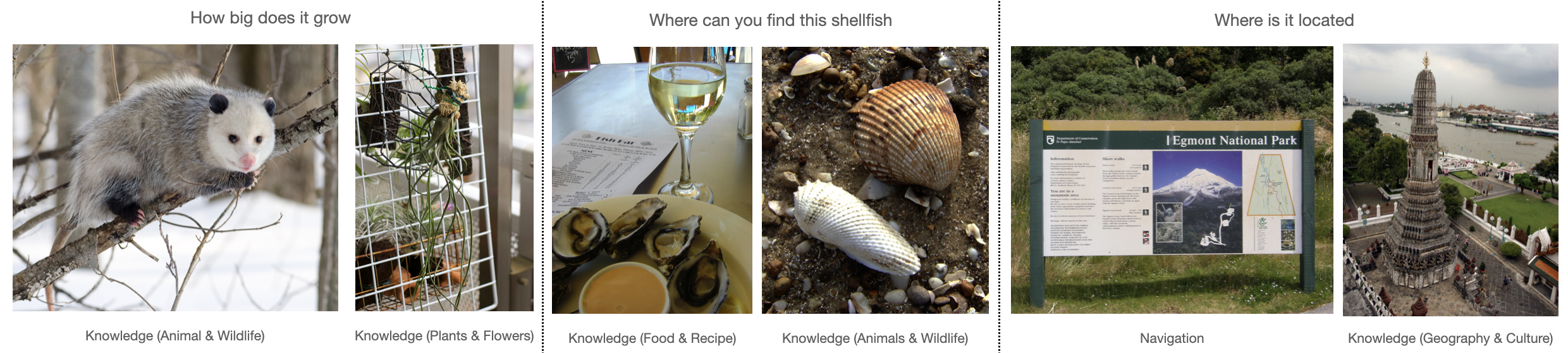}
  \caption{Examples of same utterance with different intents based on image. (a) 'How big does it grow' can be knowledge intent for animals when referring opossum while it can be knowledge intent for plants}
  \label{fig:same_utterance_different_intent}
\end{figure*}

\begin{table*}
\renewcommand\thetable{2}
\centering
\label{tab:qualitative_analysis}
\begin{adjustbox}{width=1\textwidth}
\begin{tabular}{r|lllll}
Images & \includegraphics[width=0.35\textwidth]{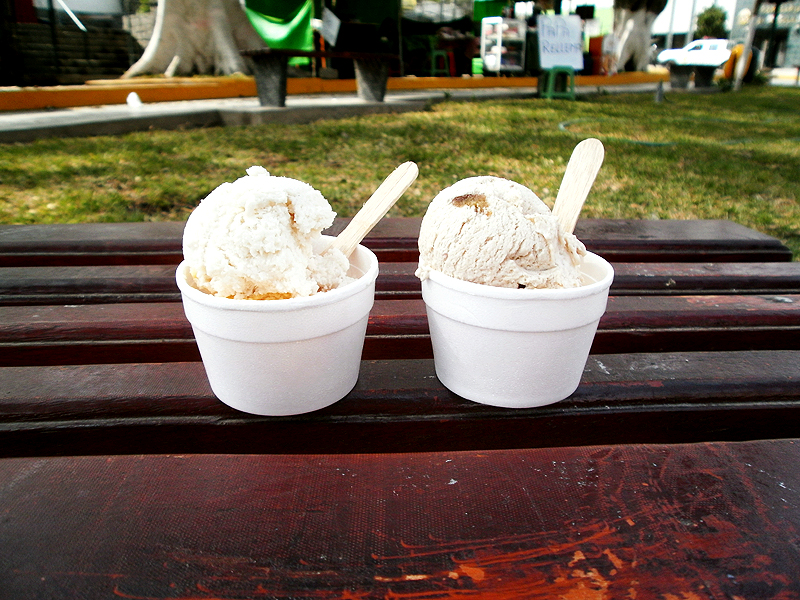} & \includegraphics[width=0.35\textwidth]{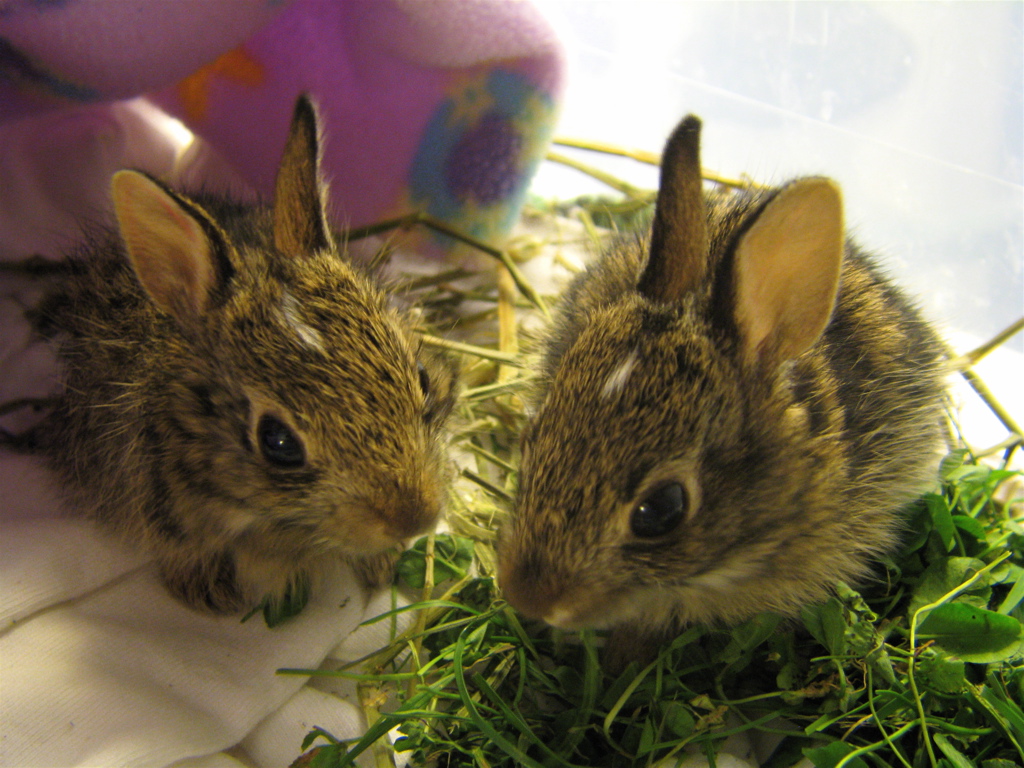} & \includegraphics[width=0.35\textwidth]{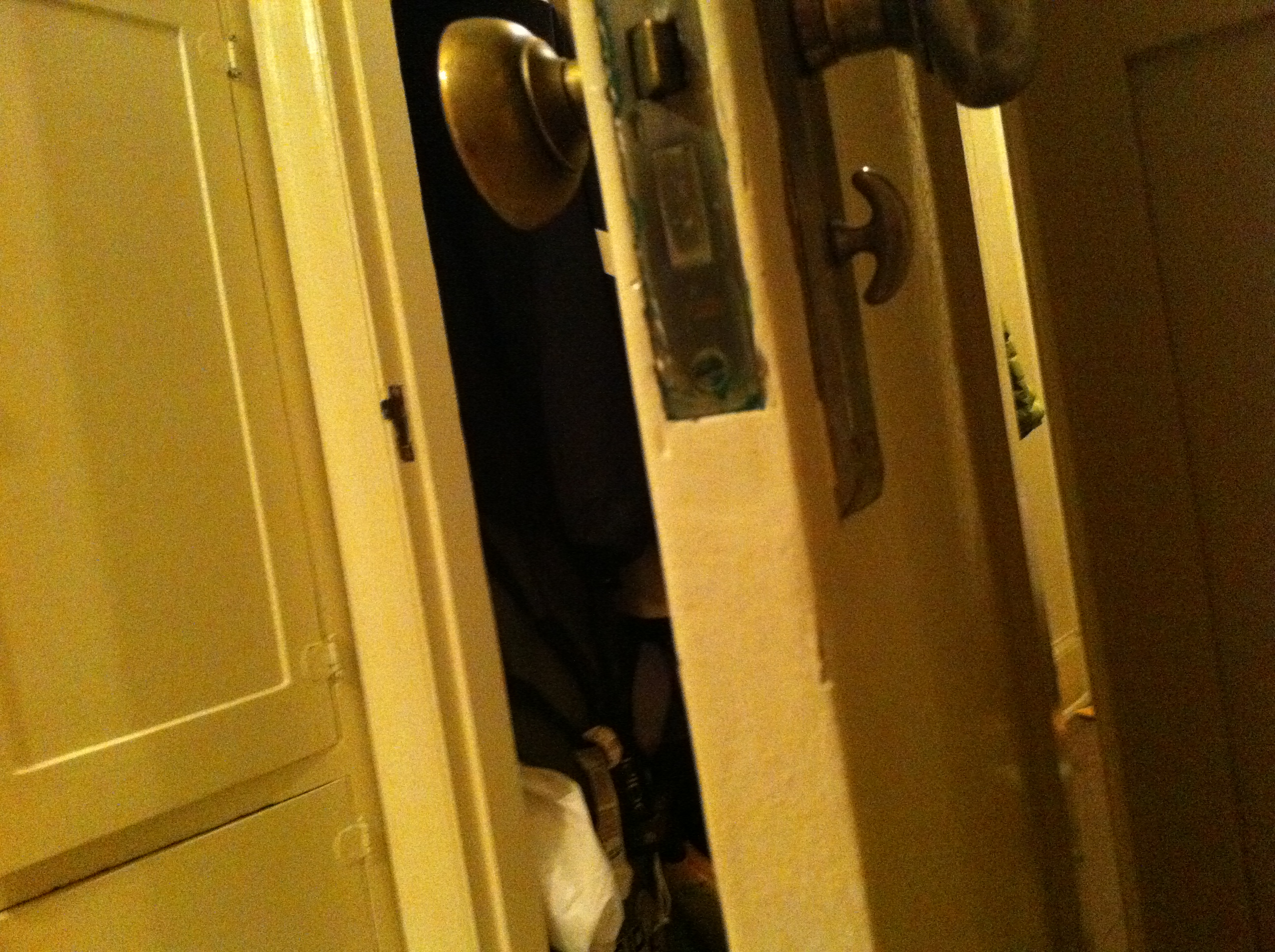}& \includegraphics[width=0.39\textwidth]{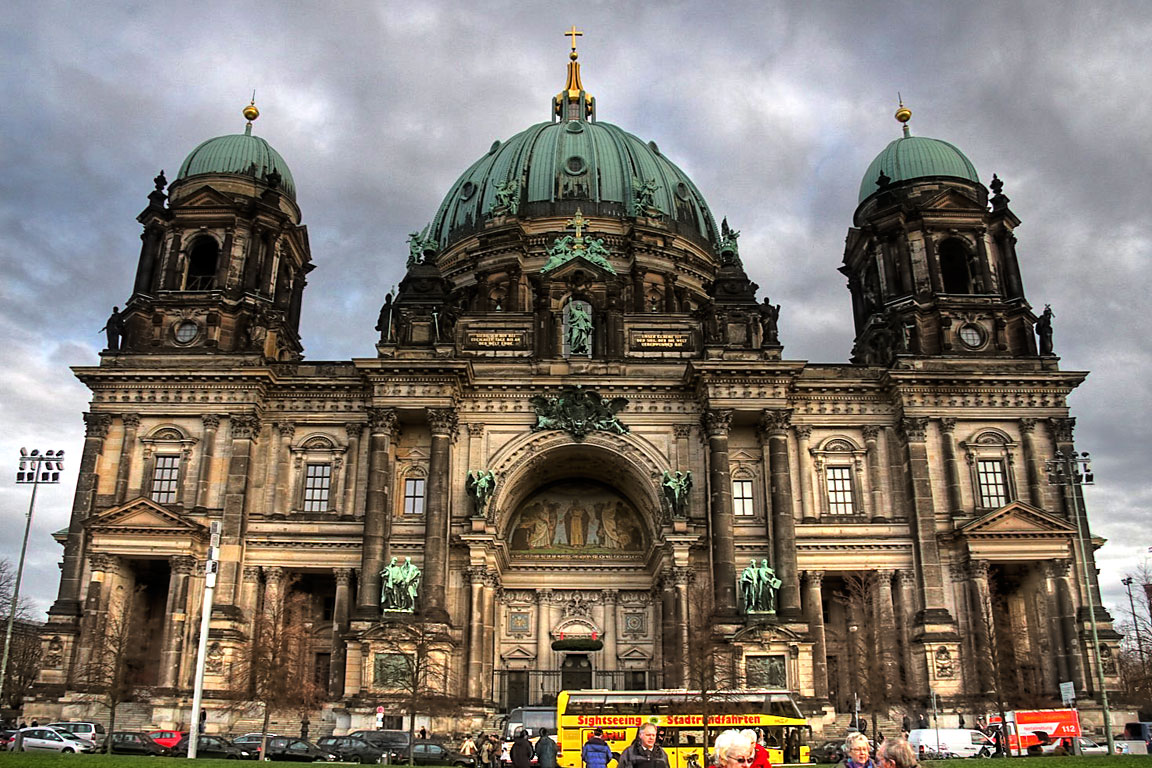}   & \includegraphics[width=0.35\textwidth]{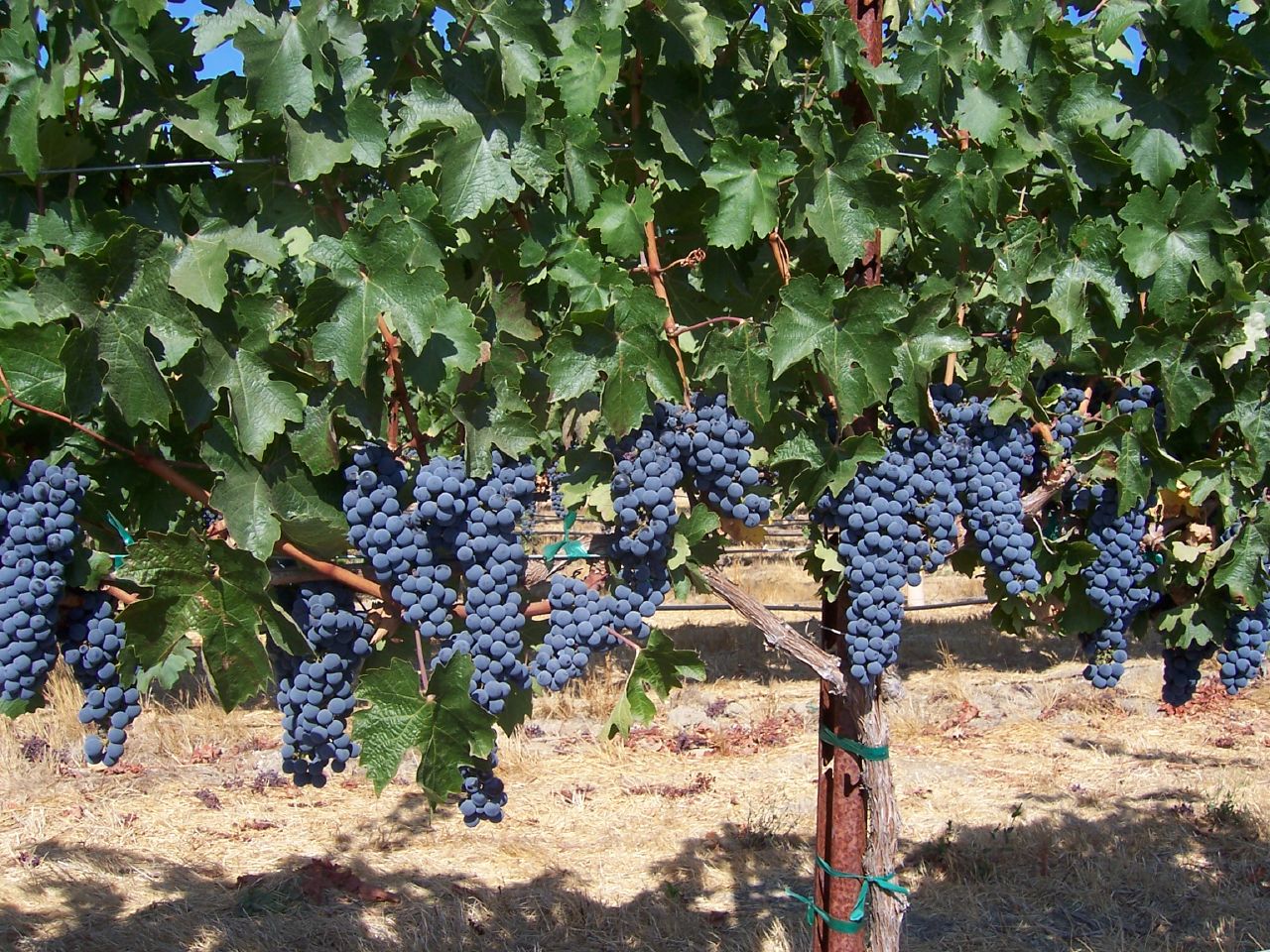}  \\
& (a) & (b) & (c) & (d)  & (e)\\
\hline
Questions &  Where could I buy this & Does this type of rabbit like carrots & What kind of lock does this door have & Is there any food for sale inside the building & Where do these kind of grapes grow \\ 
Ground Truth Intent &  {\color{darkgreen} Local Business Info Search} &  {\color{darkgreen}Knowledge (Animals \& Wildlife)} &  {\color{darkgreen}Knowledge (Other Objects)} &  {\color{darkgreen}Local Business Info Search} &  {\color{darkgreen}Knowledge (Plants \& Flowers)} \\  
\hline
Text Only  &  {\color{blue} Local Business Info Search} & Knowledge (Food \& Recipes) & Knowledge (Geography \& Culture) & Knowledge (Geography \& Culture)  &  Knowledge (Food \& Recipes) \\ 
Image Only &  Knowledge (Food \& Recipes) & {\color{blue} Knowledge (Animals \& Wildlife)} &  Knowledge (Geography \& Culture) & Knowledge (Geography \& Culture)  & Knowledge (Food \& Recipes) \\ 
Early Fusion &  Knowledge (Food \& Recipes) & Knowledge (Animals \& Wildlife) &   {\color{blue}Knowledge (Other Objects)} & Knowledge (Geography \& Culture) & Knowledge (Food \& Recipes) \\ 
Late Fusion & Knowledge (Food \& Recipes) & Knowledge (Animals \& Wildlife) &  Knowledge (Geography \& Culture) &  {\color{blue}Local Business Info Search} & Knowledge (Food \& Recipes) \\ 
LXMERT (no fine tune) &  Knowledge (Food \& Recipes) & Knowledge (Animals \& Wildlife) & Knowledge (Geography \& Culture) & Local Business Info Search &  {\color{blue}Knowledge (Plants \& Flowers)} \\  \cline{1-6}
\end{tabular}
\end{adjustbox}
\caption{Qualitative analysis of various methods for intent classification. The blue colored prediction indicates that respective strategy could predict the correct intent while others could not get the prediction right}
\vspace{-0.12cm}
\end{table*}
\section*{Abstract}
\begin{figure}[H]
\renewcommand\thefigure{1}
\begin{adjustbox}{width=\columnwidth,center}
 \includegraphics{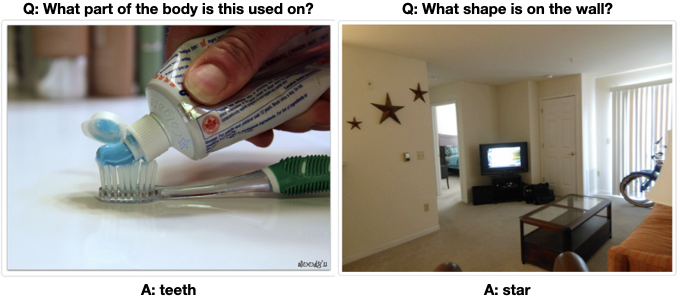}
\end{adjustbox}
 \caption{Some selected questions from the images provided in OK-VQA. The questions shown along with images may not require help from digital assistant for a visually-abled person as the answers seem obvious.}
 \label{fig:dataset_diff}
\end{figure}

 Multimodal assistants leverage vision as an additional input signal along with other modalities. However, the identification of user intent becomes a challenging task as the visual input might influence the outcome. Current digital assistants rely on spoken input and try to determine the user intent from conversational or device context. However, a dataset which includes visual input (i.e., images or videos) corresponding to questions targeted for multimodal assistant use cases is not readily available. While work in visual question answering (VQA) and visual question generation (VQG) is an important step forward, this research does not capture questions that a visually-abled person would ask multimodal assistants.  Moreover, several questions do not seek information from external knowledge \cite{Jain_CVPR2017, Mostafazadeh_ACL2016}. Recently, the OK-VQA dataset \cite{Marino2019}  tries to address this shortcoming by including questions that need to reason over unstructured knowledge. However, we make two main observations about its unsuitability for multimodal assistant use cases.  Firstly, the image types in OK-VQA datasets are often not appropriate to allow meaningful questions to be posed to the digital assistant. Secondly, the OK-VQA dataset has many obvious or common-sense questions pertaining to its images, as shown in Fig. \ref{fig:dataset_diff}, which are not challenging enough to ask a digital assistant. 

The task of identifying the intent in the given question could be challenging because of the ambiguity that can arise due to the visual context in the image. For example, as shown in the Fig. \ref{fig:same_utterance_different_intent}, the same question can have different intents based on the visual contents. Thus, the intent understanding process must take into account both the question and the image to correctly identify the intent. Various techniques have been proposed to combine textual and visual features to perform joint understanding. These approaches mainly use either fusion-based methods to combine the independently learned features for both the modalities and then use this joint representation for a given task, or use attention-based methods where a joint representation is learned by attending to relevant parts of the modalities simultaneously \cite{Nguyen2018, Tan2019, Lu2019, Chen2020}. We provide comprehensive experiments of various image and text representation strategies and its effect on intent classification.

To address the dataset issue, we introduce an effective dataset of images and corresponding natural questions that are more suitable for a multimodal assistant. To the best of our knowledge, this dataset is the first of its kind. We call it, the MMIU (Multi-modal Intent Understanding) dataset. We collected about 12K images and asked annotators to come up with questions that they would ask a multimodal assistant. Thus, we obtained 44K questions for 12K images, ensuring their applicability to digital assistants. We then created an annotation task where given an (image, question) pair, the annotators provide the underlying intent. Based on the nature of data, we pre-determined 14 different intent classes for annotators to choose from. Our dataset includes questions for factoid/descriptive information, searching for local business, asking for the recipe of food items, navigating to a specific address, chit-chat conversation about visual contents, and translating observed foreign text into the target language.

We then build a multi-class classification model that leverages visual features from the image and textual features from the question to classify a given (image, question) pair into 1 of 14 intents. To understand the effect of visual features, we use pre-trained CNNs such as VGG19 \cite{Simonyan2015}, ResNet152 \cite{He2016}, DenseNet161 \cite{Huang2017}, Inceptionv3 \cite{Szegedy2014}, MobileNetv2 \cite{Sandler2018} to get the image representation. We also experiment with recent vision transformers such as ViT \cite{Dosovitskiy2020} to see if they do any better at this task compared to traditional CNNs.  To understand the role of textual features derived from the question, we use popular transformer-based text representation strategies such as BERT \cite{Devlin2018}, RoBERTa \cite{Liu2019}, ALBERT \cite{Lan2019}, DistillBERT \cite{Sanh2019} to get the contextual representation of the question. We also experiment with combining these two modalities using early and late fusion approaches to see the overall effect on performance. Finally, we leverage a few state-of-the-art multimodal transformers such as ViLBERT \cite{Lu2019}, VL-BERT \cite{WeijieSuXizhouZhuYueCaoBinLiLeweiLuFuruWei2020}, LXMERT \cite{Tan2019}, UNITER \cite{Chen2019} which have shown impressive results on various vision and language tasks, to check their effect on our intent classification task.

We use standard evaluation metrics commonly used for multi-class classification \cite{Grandini2020}. The results of some selected experiments are shown in Table 1. In our early results, we notice that the use of text-only features dominates the intent classification task. However, the best weighted-F1 score with DistillBERT is far from ideal. The results from the fusion approaches indicate that the vanilla fusion methods are not effectively leveraging the image modality during classification. Moreover, leveraging off-the-shelf multimodal transformers such as LXMERT does not seem to help


 \begin{table}[h]
 \renewcommand\thetable{1}
 \centering
 \begin{adjustbox}{width=\columnwidth,center}
 \begin{tabular}{ c | c c c }
 \toprule
 \textbf{Strategy} & \textbf{Micro-F1}  & \textbf{Macro-F1}    &\textbf{Weighted-F1}   \\
 \hline
 Text Only (DistillBERT) & 0.7389 & 0.6519 & 0.7295 \\
 Image Only (VGG19)  & 0.3152 & 0.3007 & 0.2982 \\
 Image Only (ViT) & 0.3290 & 0.2124 & 0.2405 \\
 \hline
 \begin{tabular}{@{}c@{}}Image  + Text, Early Fusion \\ (VGG19 + DistillBERT) \end{tabular}  & 0.7342 & 0.6674 & 0.7268 \\
 \begin{tabular}{@{}c@{}}Image + Text, Late Fusion \\ (VGG19 + DistillBERT) \end{tabular}  & 0.7366 & 0.6734 & 0.7282  \\
 LXMERT (fine-tuned)  & 0.6792 & 0.6163 & 0.6726  \\ \cline{1-4}
 \bottomrule
 \end{tabular}
 \end{adjustbox}
 \caption{Results of fusion and multimodal transformer based approaches}
 \label{tab:fusion_transformer}
 \end{table}


\noindent much either.  Our qualitative analysis suggests that there is potential to leverage the best of both worlds as shown in Table 2. Thus, we need a better model architecture that combines the visual and language features more efficiently. We provide a benchmark on the newly formed MMIU dataset and plan to make it public. We hope that this dataset and the accompanying baseline results will open up new possibilities of research in the multimodal digital assistant space for the research community.

\bibliographystyle{acl_natbib}
\bibliography{custom}

\begin{thebibliography}{20}
\expandafter\ifx\csname natexlab\endcsname\relax\def\natexlab#1{#1}\fi

\bibitem[{Chen et~al.(2020)Chen, Li, Yu, {El Kholy}, Ahmed, Gan, Cheng, and
  Liu}]{Chen2020}
Yen~Chun Chen, Linjie Li, Licheng Yu, Ahmed {El Kholy}, Faisal Ahmed, Zhe Gan,
  Yu~Cheng, and Jingjing Liu. 2020.
\newblock \href {https://doi.org/10.1007/978-3-030-58577-8_7} {{UNITER:
  UNiversal Image-TExt Representation Learning}}.
\newblock In \emph{Lecture Notes in Computer Science (including subseries
  Lecture Notes in Artificial Intelligence and Lecture Notes in
  Bioinformatics)}.

\bibitem[{Chen et~al.(2019)Chen, Li, Yu, Kholy, Ahmed, Gan, Cheng, and
  Liu}]{Chen2019}
Yen-Chun Chen, Linjie Li, Licheng Yu, Ahmed~El Kholy, Faisal Ahmed, Zhe Gan,
  Yu~Cheng, and Jingjing Liu. 2019.
\newblock \href {http://arxiv.org/abs/1909.11740} {{UNITER: Learning UNiversal
  Image-TExt Representations}}.
\newblock pages 1--13.

\bibitem[{Devlin et~al.(2018)Devlin, Chang, Lee, and Toutanova}]{Devlin2018}
Jacob Devlin, Ming-Wei Chang, Kenton Lee, and Kristina Toutanova. 2018.
\newblock \href {https://doi.org/arXiv:1811.03600v2} {{BERT: Pre-training of
  Deep Bidirectional Transformers for Language Understanding}}.
\newblock \emph{arXiv:1810.04805 [cs]}.

\bibitem[{Dosovitskiy et~al.(2020)Dosovitskiy, Beyer, Kolesnikov, Weissenborn,
  Zhai, Unterthiner, Dehghani, Minderer, Heigold, Gelly, Uszkoreit, and
  Houlsby}]{Dosovitskiy2020}
Alexey Dosovitskiy, Lucas Beyer, Alexander Kolesnikov, Dirk Weissenborn,
  Xiaohua Zhai, Thomas Unterthiner, Mostafa Dehghani, Matthias Minderer, Georg
  Heigold, Sylvain Gelly, Jakob Uszkoreit, and Neil Houlsby. 2020.
\newblock \href {http://arxiv.org/abs/2010.11929} {{An Image is Worth 16x16
  Words: Transformers for Image Recognition at Scale}}.

\bibitem[{Grandini et~al.(2020)Grandini, Bagli, and Visani}]{Grandini2020}
Margherita Grandini, Enrico Bagli, and Giorgio Visani. 2020.
\newblock \href {http://arxiv.org/abs/2008.05756} {{Metrics for Multi-Class
  Classification: an Overview}}.
\newblock pages 1--17.

\bibitem[{He et~al.(2016)He, Zhang, Ren, and Sun}]{He2016}
Kaiming He, Xiangyu Zhang, Shaoqing Ren, and Jian Sun. 2016.
\newblock \href {https://doi.org/10.1109/CVPR.2016.90} {{Deep residual learning
  for image recognition}}.
\newblock In \emph{Proceedings of the IEEE Computer Society Conference on
  Computer Vision and Pattern Recognition}, volume 2016-December, pages
  770--778.

\bibitem[{Huang et~al.(2017)Huang, Liu, {Van Der Maaten}, and
  Weinberger}]{Huang2017}
Gao Huang, Zhuang Liu, Laurens {Van Der Maaten}, and Kilian~Q. Weinberger.
  2017.
\newblock \href {https://doi.org/10.1109/CVPR.2017.243} {{Densely connected
  convolutional networks}}.
\newblock In \emph{Proceedings - 30th IEEE Conference on Computer Vision and
  Pattern Recognition, CVPR 2017}.

\bibitem[{Jain et~al.(2017)Jain, Zhang, and Schwing}]{Jain_CVPR2017}
Unnat Jain, Ziyu Zhang, and Alexander Schwing. 2017.
\newblock \href {https://doi.org/10.1109/CVPR.2017.575} {{Creativity:
  Generating Diverse Questions using Variational Autoencoders}}.
\newblock \emph{Proceedings - 30th IEEE Conference on Computer Vision and
  Pattern Recognition, CVPR 2017}.

\bibitem[{Lan et~al.(2019)Lan, Chen, Goodman, Gimpel, Sharma, and
  Soricut}]{Lan2019}
Zhenzhong Lan, Mingda Chen, Sebastian Goodman, Kevin Gimpel, Piyush Sharma, and
  Radu Soricut. 2019.
\newblock {Albert: A lite bert for self-supervised learning of language
  representations}.

\bibitem[{Liu et~al.(2019)Liu, Ott, Goyal, Du, Joshi, Chen, Levy, Lewis,
  Zettlemoyer, and Stoyanov}]{Liu2019}
Yinhan Liu, Myle Ott, Naman Goyal, Jingfei Du, Mandar Joshi, Danqi Chen, Omer
  Levy, Mike Lewis, Luke Zettlemoyer, and Veselin Stoyanov. 2019.
\newblock \href {http://arxiv.org/abs/1907.11692} {{RoBERTa: A robustly
  optimized BERT pretraining approach}}.
\newblock \emph{arXiv}, (1).

\bibitem[{Lu et~al.(2019)Lu, Batra, Parikh, and Lee}]{Lu2019}
Jiasen Lu, Dhruv Batra, Devi Parikh, and Stefan Lee. 2019.
\newblock \href {http://arxiv.org/abs/1908.02265} {{ViLBERT: Pretraining
  Task-Agnostic Visiolinguistic Representations for Vision-and-Language
  Tasks}}.
\newblock (NeurIPS):1--11.

\bibitem[{Marino et~al.(2019)Marino, Rastegari, Farhadi, and
  Mottaghi}]{Marino2019}
Kenneth Marino, Mohammad Rastegari, Ali Farhadi, and Roozbeh Mottaghi. 2019.
\newblock \href {http://arxiv.org/abs/1906.00067} {{OK-VQA: A Visual Question
  Answering Benchmark Requiring External Knowledge}}.

\bibitem[{Mostafazadeh et~al.(2016)Mostafazadeh, Misra, Devlin, Mitchell, He,
  and Vanderwende}]{Mostafazadeh_ACL2016}
Nasrin Mostafazadeh, Ishan Misra, Jacob Devlin, Margaret Mitchell, Xiaodong He,
  and Lucy Vanderwende. 2016.
\newblock \href {https://doi.org/10.18653/v1/p16-1170} {{Generating natural
  questions about an image}}.
\newblock \emph{54th Annual Meeting of the Association for Computational
  Linguistics, ACL 2016 - Long Papers}, 3:1802--1813.

\bibitem[{Nguyen and Okatani(2018)}]{Nguyen2018}
Duy~Kien Nguyen and Takayuki Okatani. 2018.
\newblock \href {https://doi.org/10.1109/CVPR.2018.00637} {{Improved Fusion of
  Visual and Language Representations by Dense Symmetric Co-attention for
  Visual Question Answering}}.
\newblock In \emph{Proceedings of the IEEE Computer Society Conference on
  Computer Vision and Pattern Recognition}.

\bibitem[{Sandler et~al.(2018)Sandler, Howard, Zhu, Zhmoginov, and
  Chen}]{Sandler2018}
Mark Sandler, Andrew Howard, Menglong Zhu, Andrey Zhmoginov, and Liang~Chieh
  Chen. 2018.
\newblock \href {https://doi.org/10.1109/CVPR.2018.00474} {{MobileNetV2:
  Inverted Residuals and Linear Bottlenecks}}.
\newblock In \emph{Proceedings of the IEEE Computer Society Conference on
  Computer Vision and Pattern Recognition}.

\bibitem[{Sanh et~al.(2019)Sanh, Debut, Chaumond, and Wolf}]{Sanh2019}
Victor Sanh, Lysandre Debut, Julien Chaumond, and Thomas Wolf. 2019.
\newblock {DistilBERT, a distilled version of BERT: Smaller, faster, cheaper
  and lighter}.

\bibitem[{Simonyan and Zisserman(2015)}]{Simonyan2015}
Karen Simonyan and Andrew Zisserman. 2015.
\newblock \href {http://arxiv.org/abs/1409.1556} {{Very deep convolutional
  networks for large-scale image recognition}}.
\newblock In \emph{3rd International Conference on Learning Representations,
  ICLR 2015 - Conference Track Proceedings}.

\bibitem[{Szegedy et~al.(2014)Szegedy, Vincent, and Ioffe}]{Szegedy2014}
Christian Szegedy, Vanhoucke Vincent, and Sergey Ioffe. 2014.
\newblock {Inception-v3:Rethinking the Inception Architecture for Computer
  Vision Christian}.
\newblock \emph{HARMO 2014 - 16th International Conference on Harmonisation
  within Atmospheric Dispersion Modelling for Regulatory Purposes,
  Proceedings}.

\bibitem[{Tan and Bansal(2019)}]{Tan2019}
Hao Tan and Mohit Bansal. 2019.
\newblock \href {https://doi.org/10.18653/v1/d19-1514} {{LXMERT: Learning
  Cross-Modality Encoder Representations from Transformers}}.

\bibitem[{{Weijie Su, Xizhou Zhu, Yue Cao, Bin Li, Lewei Lu, Furu
  Wei}(2020)}]{WeijieSuXizhouZhuYueCaoBinLiLeweiLuFuruWei2020}
Jifeng~Dai {Weijie Su, Xizhou Zhu, Yue Cao, Bin Li, Lewei Lu, Furu Wei}. 2020.
\newblock \href {http://arxiv.org/abs/1908.08530v4} {{VL-BERT: Pre-training of
  Generic Visual-Linguistic Representations}}.
\newblock pages 1--16.

\end{thebibliography}

\end{document}